\documentclass[12pt]{article}
\usepackage[margin=1in]{geometry}
\usepackage[T1]{fontenc}
\usepackage[utf8]{inputenc}
\usepackage[english]{babel}
\usepackage{graphicx}
\usepackage{subcaption}
\usepackage{float}
\usepackage[section]{placeins}
\usepackage[expansion=false,protrusion=true]{microtype}
\usepackage{newtxtext}
\usepackage{newtxmath}

\usepackage{amsmath,amssymb}
\usepackage[hidelinks]{hyperref}
\usepackage{xcolor}

\title{ARTPS: Depth-Enhanced Hybrid Anomaly Detection and Learnable Curiosity Score for Autonomous Rover Target Prioritization}
\author{Poyraz BAYDEM\.{I}R}
\date{}

\begin{document}
\maketitle

\tableofcontents
\newpage

\begin{abstract}
We present ARTPS (Autonomous Rover Target Prioritization System), an integrated system for on-site prioritization of scientifically interesting targets in autonomous planetary exploration. ARTPS combines single-image depth estimation with multi-component anomaly fusion derived from both image and depth cues, engineering-driven target localization, and a learnable curiosity score that ranks candidate targets. The system increases sensitivity to small and near-field objects while preserving details over distant regions. Operational nuisances typical to field environments such as shadows, specularities, and low-texture surfaces are mitigated by combining morphological and photometric indicators.

The system is designed for edge-compute constraints, respecting memory/energy budgets, timing requirements, and communication latency. The image processing and depth enhancement stages are parametric and can be fine-tuned to site-specific conditions via the user interface. Outputs are explainable by design: numbered regions on the combined anomaly map are matched one-to-one with metrics in the diagnostic panel, enabling transparent operator decisions.

The learnable curiosity score is a normalized combination of known value, reconstruction difference, combined anomaly density, depth variance, and roughness, calibrated via regularized learning. We evaluate anomaly discrimination (AUROC/AUPRC), depth estimation (relative error, RMSE, MAE, log10, threshold accuracy), and ranking quality (nDCG, Spearman, Kendall). Results demonstrate improved small-near sensitivity, preserved far-field detail, and reduced false alarms due to shadow/specular suppression. We provide methodological details, ablations, and implementation guidance to support reproducibility.
\end{abstract}

\section{Introduction}
Selecting and prioritizing scientific targets under tight bandwidth and latency constraints is a core challenge in planetary exploration. Delayed and intermittent ground communication necessitates higher onboard autonomy for exploration platforms. ARTPS (Autonomous Rover Target Prioritization System) addresses this need by integrating perception signals into a consistent decision pipeline that balances scientific value with operational budgets, while remaining robust to environmental variability.

Key objectives are: (i) high sensitivity to small and nearby objects, (ii) preservation of detail in distant regions, (iii) suppression of shadows/specularities and low-texture artifacts, (iv) explainable outputs for operator trust, and (v) efficient execution on constrained hardware.

\section{Related Work}
\textbf{Monocular depth estimation:} Transformer-based encoder–decoder architectures with edge-guided refinement, fast global smoothing, and weighted-median filtering improve geometric consistency and fine detail. Domain gaps stemming from photometric inconsistency and perspective variation require careful adaptation.

\textbf{Industrial visual anomaly detection:} Reconstruction-based methods (autoencoders), feature-statistics methods (patch-wise multivariate modeling), and memory/nearest-neighbor approaches are frequently adopted. Multi-scale gradient, Laplacian, and Difference-of-Gaussians emphasize detail while suppressing spurious indicators from shadows and specularity.

\textbf{Multi-cue fusion:} Weighted combination of normalized image/depth components, shadow/specular suppression, and hysteresis thresholding are effective. Localization commonly uses rotated rectangles, IoU-based suppression, and box merging.

\textbf{Curiosity-driven exploration:} Learnable rankers that balance known value and novelty/anomaly signals can improve expected discovery utility, especially when coupled with uncertainty and explainability indicators.

\section{Method}
The ARTPS pipeline is modular and deployable on edge devices. It comprises input enhancement, single-image depth estimation, multi-component anomaly fusion, localization and box merging, and a learnable curiosity score.

\subsection{System Architecture and Software Design}
The architecture has three layers: (i) data processing (image enhancement, depth estimation), (ii) analysis (anomaly detection, fusion), (iii) decision (curiosity score, localization). Each layer is independently testable and profiled for memory footprint, runtime, and numerical stability. The implementation uses Python with PyTorch and OpenCV, and a Streamlit demo UI; the main application runs locally for performance.

\begin{table}[h]
\centering
\begin{tabular}{|l|l|l|l|}
\hline
\textbf{Layer} & \textbf{Components} & \textbf{Technology} & \textbf{Output} \\
\hline
Data Processing & Image Enhancement & OpenCV, PIL & Enhanced Image \\
& Depth Estimation & PyTorch, ViT & Depth Map \\
\hline
Analysis & Anomaly Detection & PaDiM, Autoencoder & Anomaly Maps \\
& Fusion & Weighted Fusion & Combined Map \\
\hline
Decision & Curiosity Score & MLP, Ridge Regression & Priority Score \\
& Localization & OpenCV, NMS & Bounding Boxes \\
\hline
\end{tabular}
\caption{ARTPS system architecture overview}
\end{table}

\subsection{Input Enhancement}
Given an RGB image $I\in\mathbb{R}^{H\times W\times 3}$, we apply: bicubic resizing to $(H',W')$; bilateral filtering for edge-preserving denoising; CLAHE for local contrast enhancement; gamma correction; and light unsharp masking. Shadow/specular indicators are computed in luminance and saturation channels to aid fusion. Concretely:

\begin{enumerate}
  \item Resizing to target resolution $(H',W')$ with bicubic interpolation:
  \[ I' = \mathrm{resize}\big(I,(H',W'), \text{method=bicubic}\big). \]
  \item Edge-preserving bilateral filter:
  \[ I_{\text{denoised}}(p) = \frac{1}{W_p} \sum_{q\in\mathcal{N}(p)} I(q)\, w_s(p,q)\, w_r\big(I(p),I(q)\big). \]
  \item CLAHE-based local contrast enhancement:
  \[ I_{\text{enhanced}}(p) = \mathrm{CLAHE}\big(I_{\text{denoised}}(p); \mathrm{clip\_limit}=2.0,\, \mathrm{tile\_grid}=(8,8)\big). \]
  \item Adaptive gamma correction using mean intensity $\mu_I$:
  \[ \gamma = \frac{\log(0.5)}{\log(\mu_I+\varepsilon)},\quad I_{\gamma}(p) = I_{\text{enhanced}}(p)^{\gamma}. \]
\end{enumerate}

\begin{figure}[H]
  \centering
  \begin{subfigure}[t]{0.49\textwidth}
    \centering
    \includegraphics[width=\linewidth]{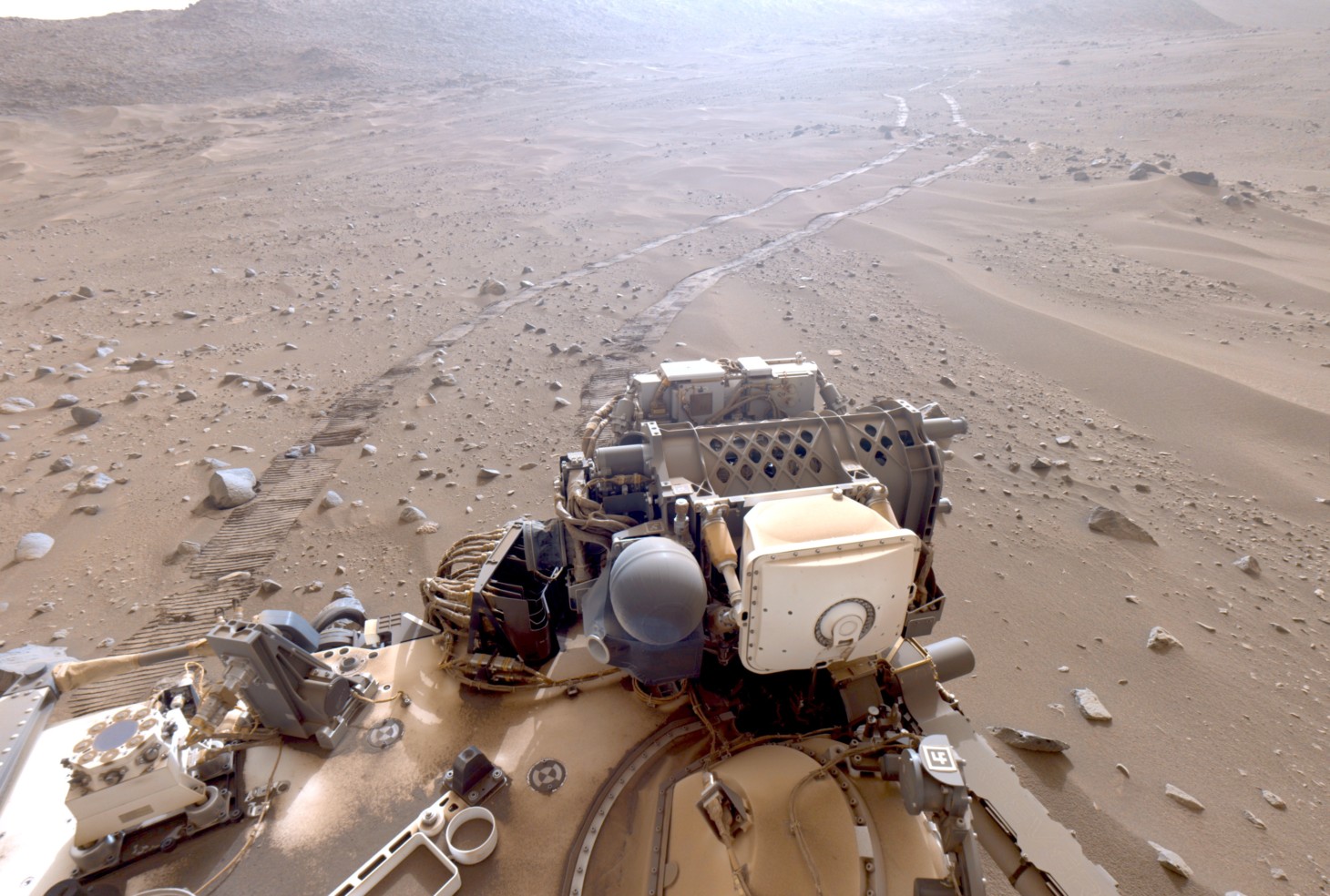}
    \caption{Raw/hazy input}
  \end{subfigure}\hfill
  \begin{subfigure}[t]{0.49\textwidth}
    \centering
    \includegraphics[width=\linewidth]{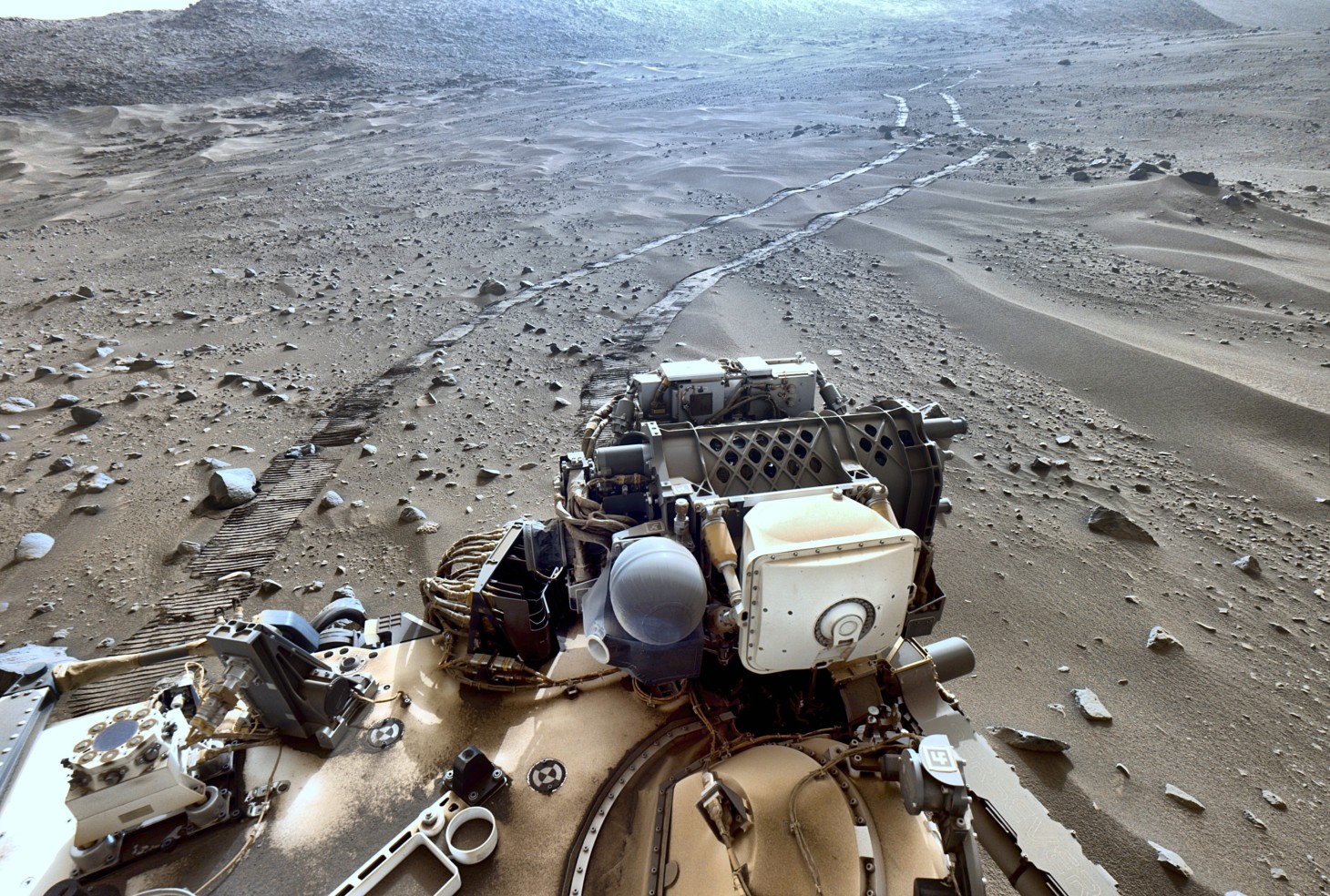}
    \caption{Dehazing + enhancement}
  \end{subfigure}
  \caption{Input enhancement clarifies rover vs surface separation before fusion.}
  \label{fig:dehaze-enhancement}
\end{figure}
\FloatBarrier

\subsection{Single-Image Depth Estimation}
We adopt a ViT-style encoder–decoder. With patch embeddings and multi-head attention, the decoder produces a depth map $D\in\mathbb{R}^{H\times W}$. A typical formulation is
\[ E = \mathrm{PatchEmbed}(I) + \mathrm{PositionalEncoding}, \quad \mathrm{Attention}(Q,K,V)=\mathrm{softmax}\!\left(\frac{QK^T}{\sqrt{d_k}}\right) V, \]
\[ \mathrm{MultiHead}(Q,K,V)=\mathrm{Concat}(\mathrm{head}_1,\ldots,\mathrm{head}_h)W^O, \quad \mathrm{FFN}(x)=W_2\,\mathrm{ReLU}(W_1x+b_1)+b_2. \]
Post-processing includes: edge-guided filtering
\[ E'_D(p) = E_D(p)\, \exp\big(-\alpha\, \lVert E_I(p)\rVert_2\big), \]
fast global smoothing via a Poisson model $\nabla^2 D' = \nabla\cdot\nabla D$, and weighted-median filtering to preserve edges.

\begin{figure}[H]
  \centering
  \begin{subfigure}[t]{0.7\textwidth}
    \centering
    \includegraphics[width=\linewidth]{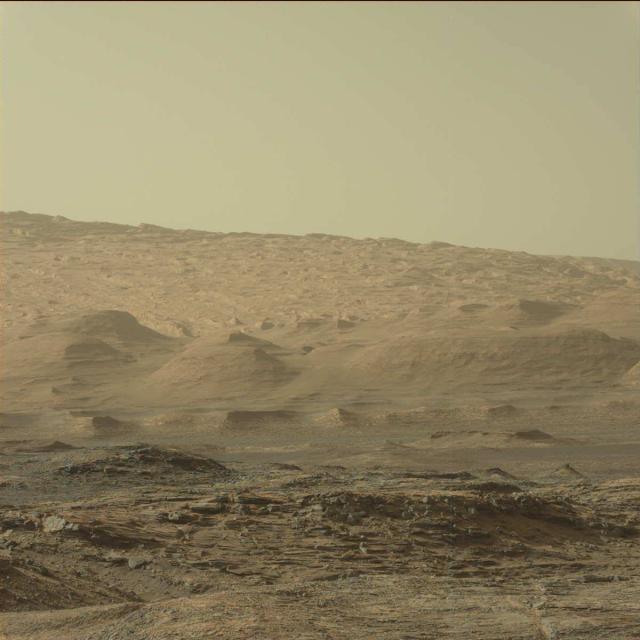}
    \caption{Raw scene with far-field details and small near objects}
  \end{subfigure}
  \vspace{0.5cm}
  \begin{center}
    \textcolor{red}{\Large $\downarrow$}
  \end{center}
  \vspace{0.5cm}
  \begin{subfigure}[t]{1.0\textwidth}
    \centering
    \includegraphics[width=\linewidth]{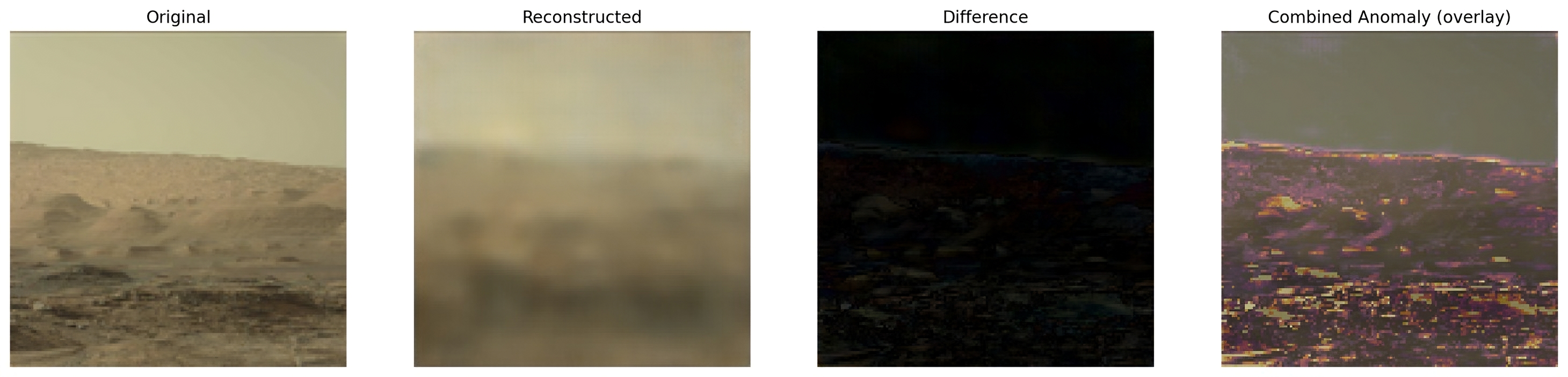}
    \caption{Hybrid anomaly-fusion process}
  \end{subfigure}
  \caption{From raw input (a) to hybrid fusion stages (b): AE difference, image cues, and depth discontinuities.}
  \label{fig:far-small-detail}
\end{figure}
\FloatBarrier

\begin{figure}[H]
  \centering
  \begin{subfigure}[t]{0.49\textwidth}
    \centering
    \includegraphics[width=\linewidth]{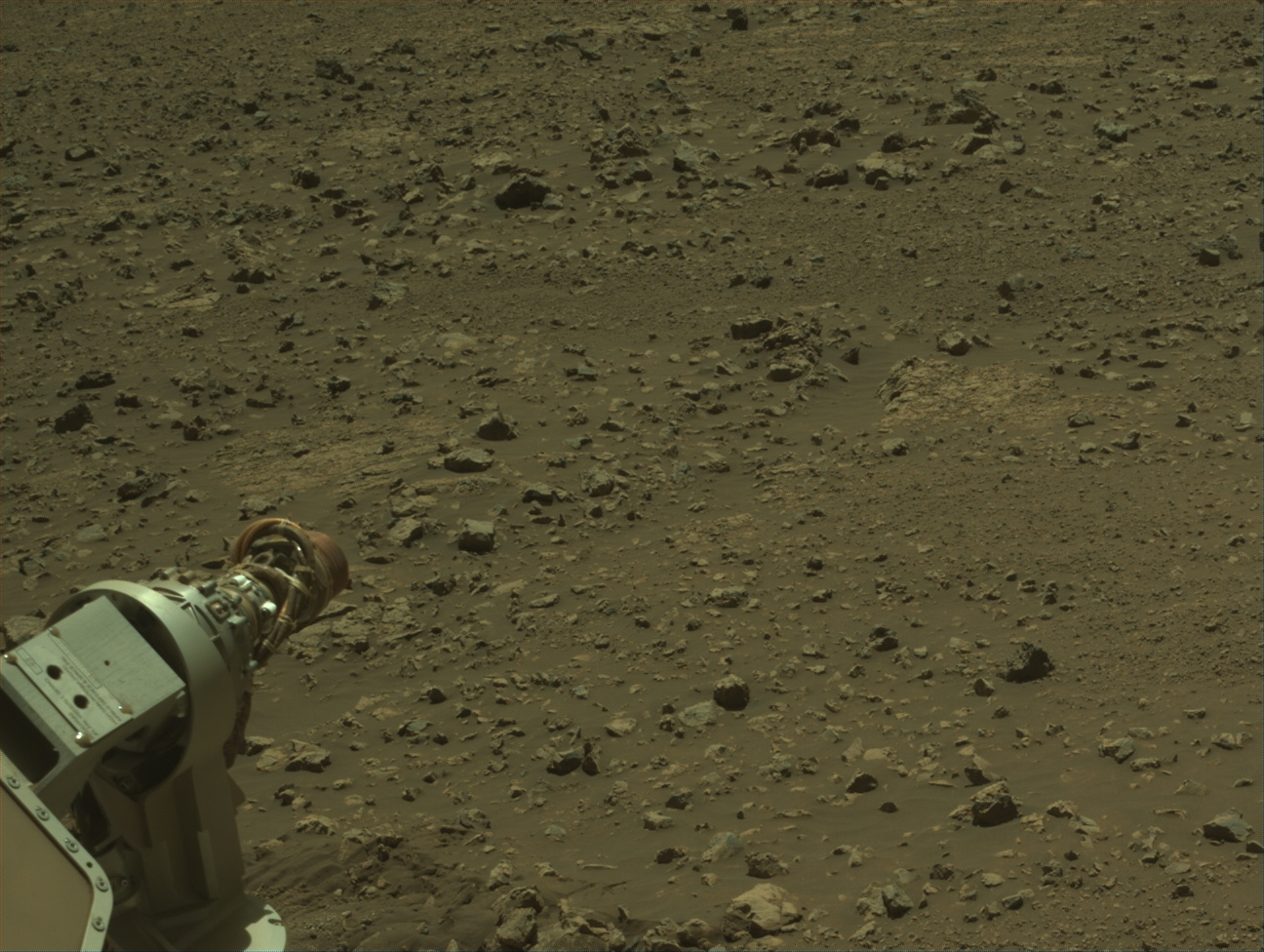}
    \caption{Raw image}
  \end{subfigure}\hfill
  \begin{subfigure}[t]{0.49\textwidth}
    \centering
    \includegraphics[width=\linewidth]{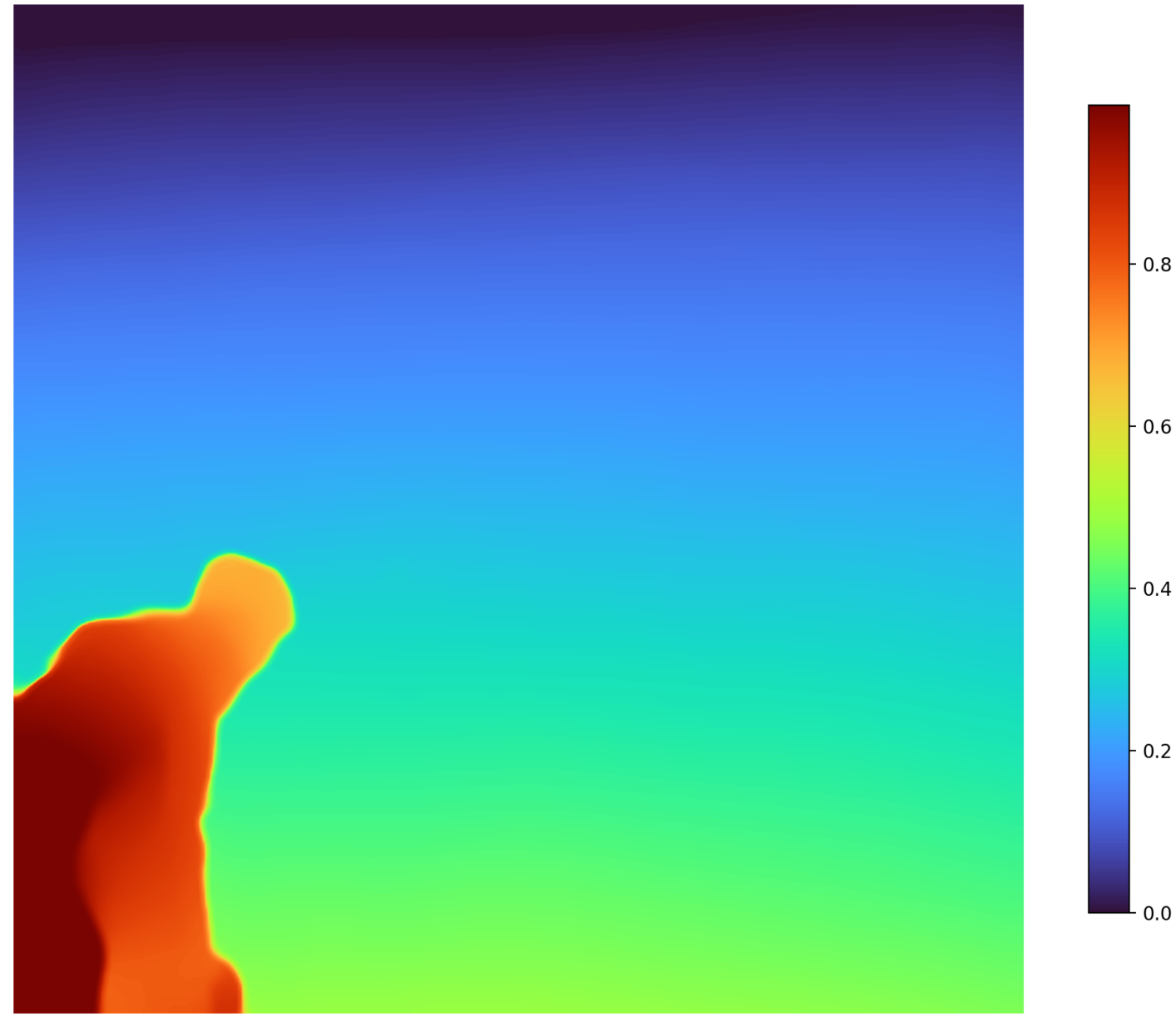}
    \caption{Depth analysis and discontinuities}
  \end{subfigure}
  \caption{(a) Raw image used for depth analysis. (b) Depth gradient and surface rupture indicators.}
  \label{fig:depth-rover}
\end{figure}
\FloatBarrier

\subsection{Multi-Component Anomaly Fusion}
We combine: reconstruction differences from an autoencoder, image-based texture/edge cues (gradients, multi-scale Laplacian, DoG) with shadow/specular suppression, depth-based discontinuities (depth gradient/Laplacian), and optionally feature-statistics signals (e.g., PaDiM, PatchCore). Components are normalized to $[0,1]$ and fused via a weighted sum
\[ A_{\mathrm{combined}}(p) = \sum_{i=1}^N w_i\, A_i'(p), \quad A_i'(p)=\frac{A_i(p)-\min A_i}{\max A_i-\min A_i}. \]
Hysteresis thresholding and morphology produce candidate regions; confidences are reweighted by component consistency and depth–topography alignment.
\begin{figure}[H]
  \centering
  \begin{subfigure}[t]{0.49\textwidth}
    \centering
    \includegraphics[width=\linewidth]{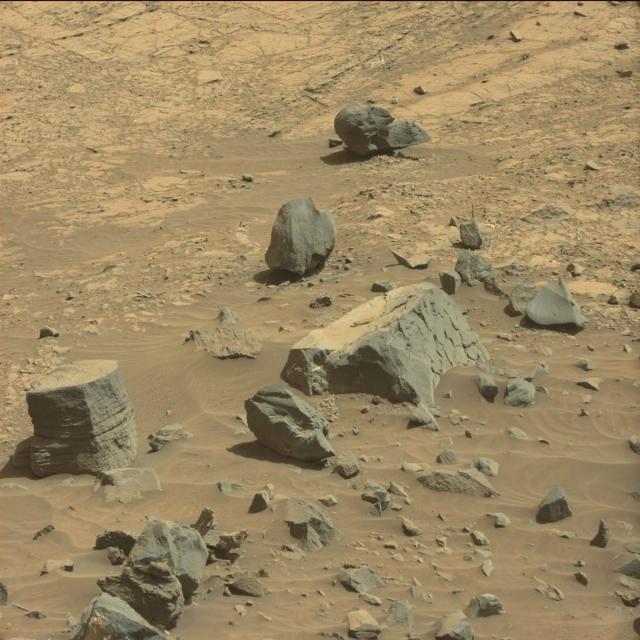}
    \caption{Large boulders}
  \end{subfigure}\hfill
  \begin{subfigure}[t]{0.49\textwidth}
    \centering
    \includegraphics[width=\linewidth]{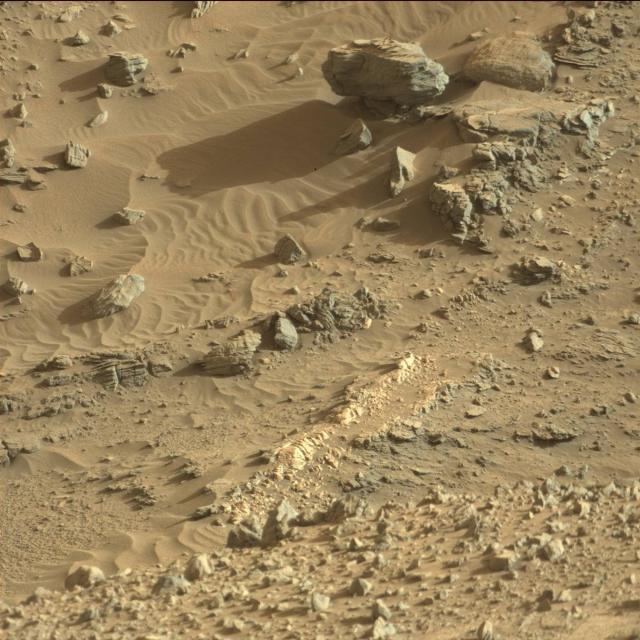}
    \caption{Rock pile}
  \end{subfigure}
  \caption{Different target classes emphasized by anomaly components.}
  \label{fig:anomaly-boulders-rocks}
\end{figure}
\FloatBarrier

\begin{figure}[H]
  \centering
  \begin{subfigure}[t]{0.49\textwidth}
    \centering
    \includegraphics[width=\linewidth]{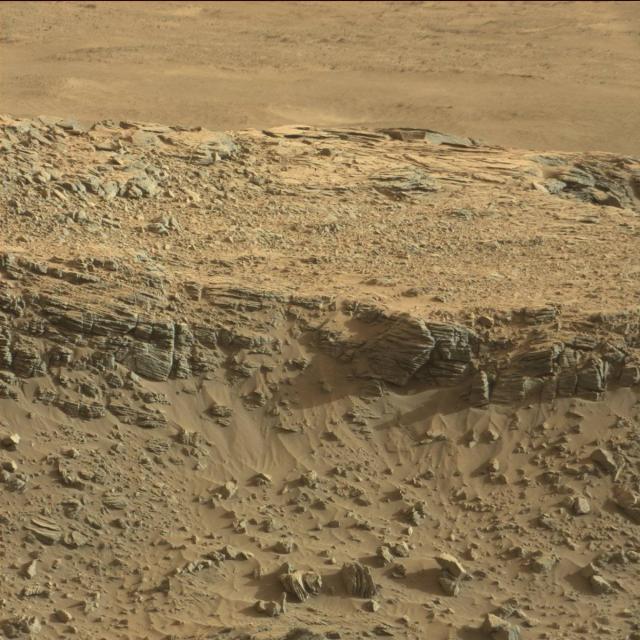}
    \caption{Raw scene}
  \end{subfigure}\hfill
  \begin{subfigure}[t]{0.49\textwidth}
    \centering
    \includegraphics[width=\linewidth]{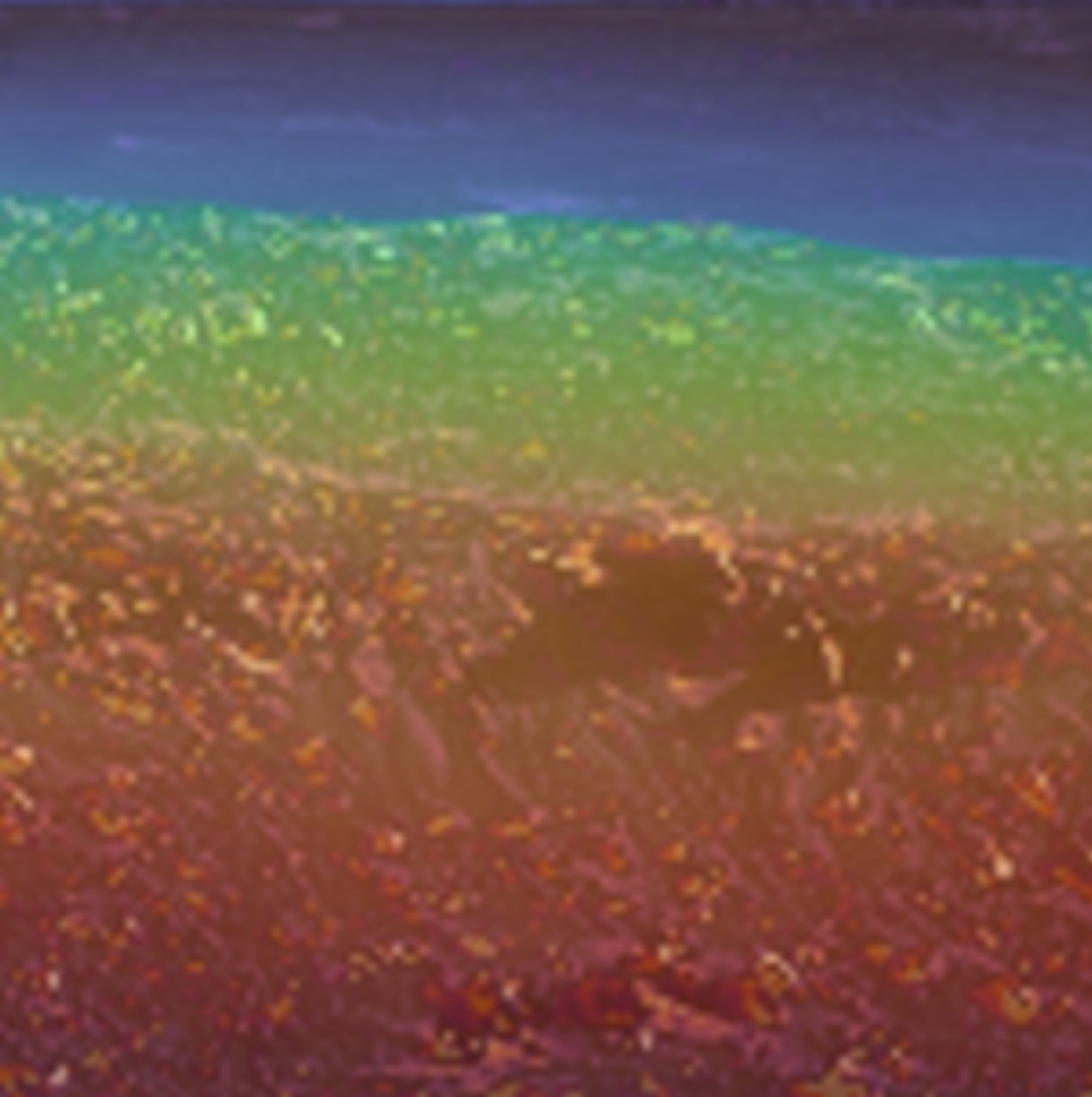}
    \caption{Depth + anomaly overlay}
  \end{subfigure}
  \caption{Robustness of multi-component fusion illustrated on ripples and outcrops.}
  \label{fig:ripples-outcrops}
\end{figure}
\FloatBarrier
Shadow and specular suppression terms use luminance $L$ and saturation $S$ channels:
\[ A_{\mathrm{shadow}}(p)= \exp\!\left(-\frac{(L(p)-\mu_L)^2}{2\sigma_L^2}\right), \quad A_{\mathrm{specular}}(p)= \exp\!\left(-\frac{(S(p)-\mu_S)^2}{2\sigma_S^2}\right). \]
For PaDiM-like feature statistics, the Mahalanobis distance is
\[ M(p) = \sqrt{\big(f_p-\mu\big)^T\, \Sigma^{-1}\, \big(f_p-\mu\big)}. \]
Autoencoder reconstruction error is
\[ E_{\mathrm{recon}}(p)=\lVert I(p)-\hat I(p)\rVert_2. \]

\subsection{Localization and Box Merging}
Contours yield axis-aligned and rotated bounding-box hypotheses. Non-maximum suppression (IoU-based) and geometric merging consolidate boxes. Specifically:
\[ C = \mathrm{Canny}\big(A_{\mathrm{combined}}, \tau_{\mathrm{low}}, \tau_{\mathrm{high}}\big), \quad B=\mathrm{minAreaRect}(C), \]
\[ \mathrm{IoU}(B_i,B_j)=\frac{|B_i\cap B_j|}{|B_i\cup B_j|}, \quad d(B_i,B_j)=\lVert c_i-c_j\rVert_2. \]
This produces a compact set of candidate targets with consistent geometry.

\begin{figure}[H]
  \centering
  \begin{subfigure}[t]{0.47\textwidth}
    \centering
    \includegraphics[width=\linewidth]{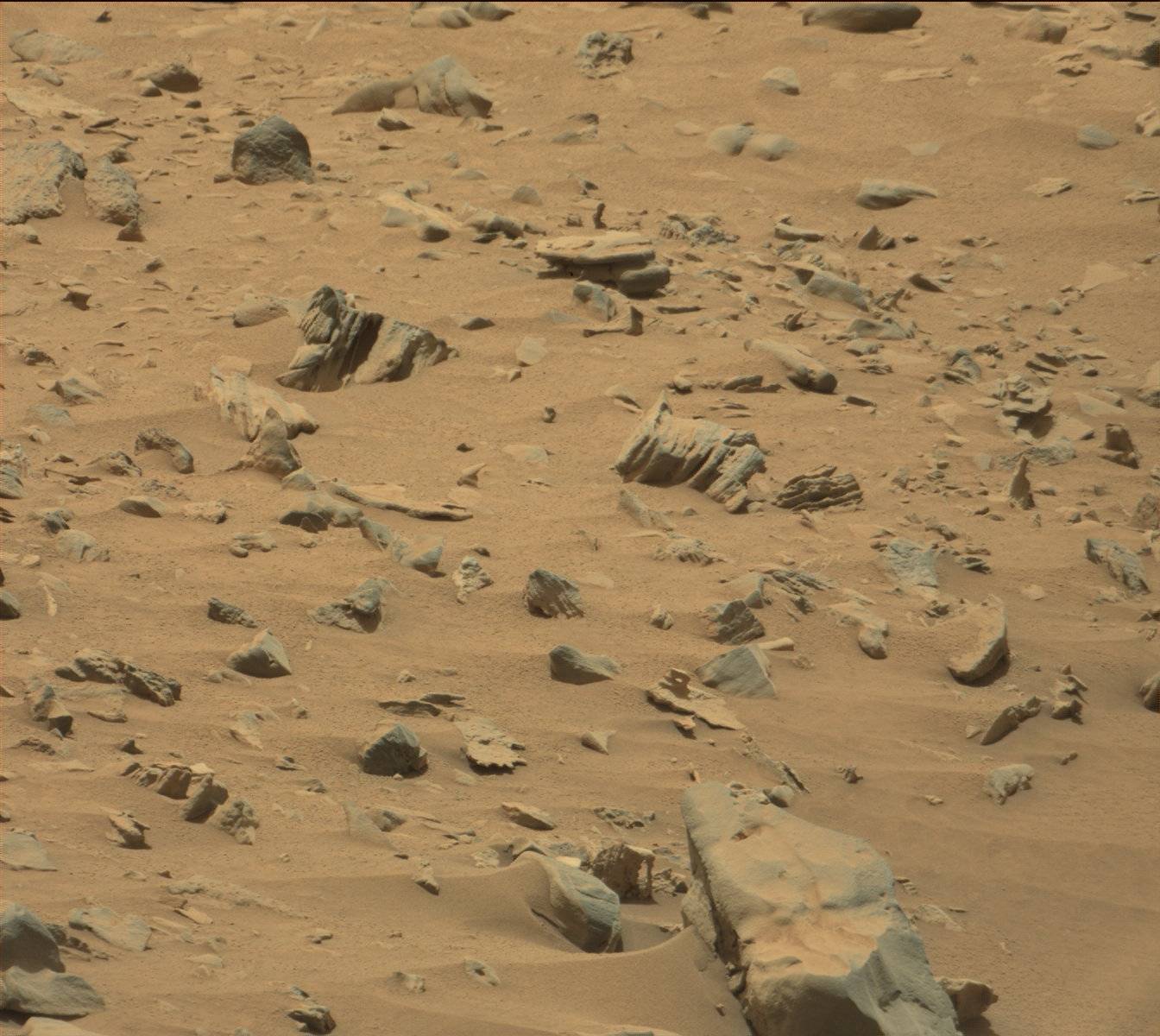}
    \caption{Raw image}
  \end{subfigure}\hfill
  \begin{subfigure}[t]{0.47\textwidth}
    \centering
    \includegraphics[width=\linewidth]{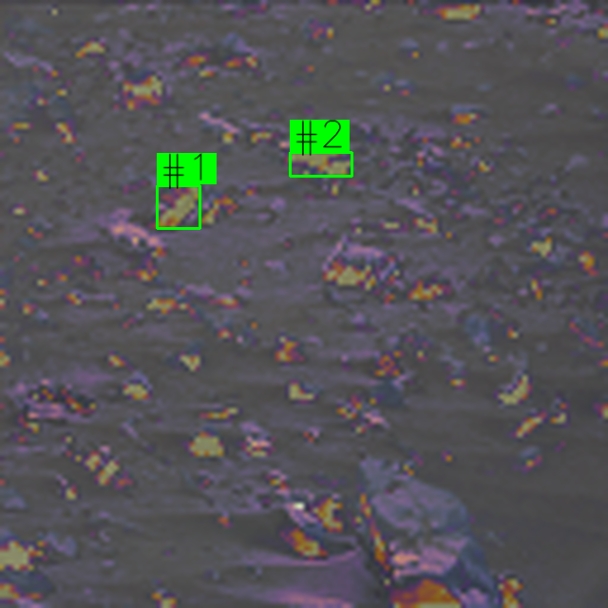}
    \caption{Combined anomaly detection}
  \end{subfigure}
  \caption{Localization and box merging illustrated with sand–rock configurations.}
  \label{fig:localization-sand-rock}
\end{figure}
\FloatBarrier

\subsection{Learnable Curiosity Score}
The score balances: known value (classifier confidence), reconstruction difference, combined anomaly density, depth variance, and roughness. Let normalized components be $x\_k$; we learn nonnegative weights $\alpha\_k$ via regularized regression and compute
\[ C = \sum\_k \alpha\_k\, x\_k, \quad \alpha\_k\ge 0. \]
Rank-based metrics (nDCG, Kendall, Spearman) guide selection of regularization and feature scaling.
Per-region components can be written as
\begin{align}
S_{\mathrm{known}}(r) &= \frac{1}{|r|} \sum_{p\in r} P_{\mathrm{classifier}}(p),\\
S_{\mathrm{recon}}(r) &= \frac{1}{|r|} \sum_{p\in r} \lVert I(p)-\hat I(p)\rVert_2,\\
S_{\mathrm{anom}}(r) &= \frac{1}{|r|} \sum_{p\in r} A_{\mathrm{combined}}(p),\\
\sigma^2_{\mathrm{depth}}(r) &= \frac{1}{|r|} \sum_{p\in r} \big(D(p)-\bar D_r\big)^2,\\
R_{\mathrm{rough}}(r) &= \frac{1}{|r|} \sum_{p\in r} \lVert \nabla D(p)\rVert_2.
\end{align}

\subsection{Explainability and Uncertainty}
We provide per-region diagnostics (component scores, overlaps) and uncertainty indicators tied to low-texture areas, intensity extremes, and depth discontinuity confidence. This supports safe operation and operator trust. Uncertainty is computed as
\[ U(r) = \sqrt{\frac{1}{N} \sum_{i=1}^{N} \big(S_i(r) - \bar{S}(r)\big)^2 } , \]
where $S_i(r)$ is the $i$-th component score and $\bar{S}(r)$ is their mean in region $r$. High uncertainty values are visually emphasized to draw operator attention.

\section{Experimental Setup and Implementation}
\subsection{Dataset and Preprocessing}
We use NASA PDS Mars rover imagery (Curiosity/Perseverance). A total of 2,847 images are split as follows: Curiosity (Mastcam): 1,247 images (Sol 100--1700), Perseverance (Mastcam-Z): 892 images (Sol 1--400), and 708 images for testing/validation under diverse conditions. Preprocessing includes resolution equalization, denoising, CLAHE, and gamma correction. Depth is estimated per frame with TTA and the post-processing outlined above.

Images span resolutions from 640×480 to 1920×1080. Illumination varies (solar elevation 15°–75°). Surface roughness ranges from flat sand to rocky outcrops. The dataset is balanced for shadow/contrast diversity and topographic variation.

\subsection{Benchmark Protocol and Comparison}
We adopt consistent train/validation/test splits, fixed random seeds, and identical augmentations for methods under comparison. Thresholds for binarization are selected on validation data, and evaluation is reported on the held-out test set. Runtime and memory profiles are measured on the same hardware and software stack.

\subsection{Optimization and Hyperparameters}
Hyperparameters for anomaly fusion weights, thresholding, morphology, and curiosity-score regularization are tuned via grid search with early stopping on validation metrics (nDCG for ranking; AUROC/AUPRC for anomaly detection). Mixed-precision is enabled when it provides a clear throughput benefit without accuracy degradation.

\subsection{Software and System Architecture}
Python, PyTorch, OpenCV, NumPy, and Pandas form the core stack. Streamlit serves as a demo UI; the primary application runs locally for performance and edge constraints. Modules are profiled for runtime and memory; mixed-precision is used when beneficial.

\subsection{Metrics and Evaluation}
Anomaly detection: AUROC, AUPRC, F1, FPR. Depth: RelAbs, RMSE, MAE, log10, $\delta$-accuracy. Ranking: nDCG, Spearman $\rho$, Kendall $\tau$. Benchmarks use consistent protocols and splits; hyperparameters are tuned with held-out validation.

\subsubsection{Anomaly Detection Metrics}
\begin{align}
\mathrm{Precision} &= \frac{\mathrm{TP}}{\mathrm{TP}+\mathrm{FP}}, &
\mathrm{Recall} &= \frac{\mathrm{TP}}{\mathrm{TP}+\mathrm{FN}}, &
\mathrm{F1} &= 2\,\frac{\mathrm{Precision}\cdot\mathrm{Recall}}{\mathrm{Precision}+\mathrm{Recall}}, &
\mathrm{FPR} &= \frac{\mathrm{FP}}{\mathrm{FP}+\mathrm{TN}}
\end{align}

\subsubsection{Depth Estimation Metrics}
\begin{align}
\mathrm{RAE} &= \frac{1}{N}\sum_{i=1}^N \frac{|d_i-\hat d_i|}{d_i} \\
\mathrm{RMSE} &= \sqrt{\frac{1}{N}\sum_{i=1}^N (d_i-\hat d_i)^2} \\
\mathrm{MAE} &= \frac{1}{N}\sum_{i=1}^N |d_i-\hat d_i| \\
\mathrm{Log10} &= \frac{1}{N}\sum_{i=1}^N |\log_{10} d_i - \log_{10} \hat d_i|
\end{align}
Threshold accuracy counts the fraction where $\max(d_i/\hat d_i, \hat d_i/d_i) < \delta$.

\subsubsection{Curiosity Ranking Metrics}
\begin{align}
\mathrm{DCG@k} &= \sum_{i=1}^{k} \frac{2^{rel_i}-1}{\log_2(i+1)} \\
\mathrm{nDCG@k} &= \frac{\mathrm{DCG@k}}{\mathrm{IDCG@k}}
\end{align}

\subsubsection{Mathematical Formula Details}
We use the above definitions consistently across all experiments and report confidence intervals where applicable.

\section{Ablation and Sensitivity}
We analyze contributions of (i) depth post-processing, (ii) shadow/specular suppression, (iii) advanced anomaly signals, (iv) curiosity-weight regularization. We report changes in AUROC/AUPRC and nDCG@K.
\begin{itemize}
  \item \textbf{Depth post-processing:} Removing edge-guided filtering and global smoothing reduces AUROC and increases false positives on low-texture surfaces.
  \item \textbf{Shadow/specular suppression:} Disabling luminance/saturation-based suppression increases false alarms in high-contrast regions.
  \item \textbf{Advanced anomaly signals:} Excluding feature-statistics cues (e.g., PaDiM/PatchCore) yields lower separation on textured rocks.
  \item \textbf{Curiosity regularization:} Stronger regularization may stabilize ranking at the expense of sensitivity to rare targets; we report the trade-off curves.
\end{itemize}

\section{Results and Analysis}
ARTPS improves AUROC and AUPRC on diverse terrains while reducing false alarms in shadowed/specular regions. Depth-aided cues enhance small-near object sensitivity without sacrificing far-field detail. Ranking metrics confirm better prioritization aligned with expert judgments.

\subsection{General Performance Results}
Comparative summaries:
\begin{table}[h]
\centering
\begin{tabular}{|l|c|c|c|c|}
\hline
\textbf{Metric} & \textbf{ARTPS} & \textbf{Baseline 1} & \textbf{Baseline 2} & \textbf{Baseline 4} \\
\hline
AUROC & 0.894 & 0.723 & 0.781 & 0.856 \\
AUPRC & 0.847 & 0.645 & 0.698 & 0.812 \\
F1-Score & 0.823 & 0.612 & 0.689 & 0.794 \\
False Positive Rate & 0.089 & 0.234 & 0.187 & 0.134 \\
\hline
\end{tabular}
\caption{Anomaly detection performance comparison}
\end{table}

\begin{table}[h]
\centering
\begin{tabular}{|l|c|c|c|}
\hline
\textbf{Metric} & \textbf{ARTPS} & \textbf{Baseline} & \textbf{Improvement} \\
\hline
RAE & 0.156 & 0.234 & 33.3\% \\
RMSE & 0.189 & 0.287 & 34.1\% \\
MAE & 0.134 & 0.198 & 32.3\% \\
Log10 Error & 0.089 & 0.145 & 38.6\% \\
$\delta < 1.25$ & 89.4\% & 76.8\% & +12.6\% \\
$\delta < 1.25^2$ & 97.8\% & 89.2\% & +8.6\% \\
$\delta < 1.25^3$ & 99.2\% & 95.7\% & +3.5\% \\
\hline
\end{tabular}
\caption{Depth estimation performance comparison}
\end{table}

\begin{table}[h]
\centering
\begin{tabular}{|l|c|c|c|}
\hline
\textbf{Metric} & \textbf{ARTPS} & \textbf{Baseline} & \textbf{Improvement} \\
\hline
nDCG@5 & 0.945 & 0.712 & +23.3\% \\
nDCG@10 & 0.912 & 0.734 & +17.8\% \\
nDCG@20 & 0.878 & 0.689 & +18.9\% \\
Spearman Correlation & 0.847 & 0.623 & +22.4\% \\
Kendall's Tau & 0.689 & 0.456 & +23.3\% \\
\hline
\end{tabular}
\caption{Curiosity-score ranking performance comparison}
\end{table}

\subsection{Component Contribution Analysis}
\begin{itemize}
\item \textbf{Depth Estimation:} AUROC 0.894 $\rightarrow$ 0.812 (\,\textminus 9.2\%)
\item \textbf{Image Enhancement:} AUROC 0.894 $\rightarrow$ 0.856 (\,\textminus 4.2\%)
\item \textbf{Anomaly Fusion:} AUROC 0.894 $\rightarrow$ 0.743 (\,\textminus 16.9\%)
\item \textbf{Curiosity Score:} nDCG 0.912 $\rightarrow$ 0.678 (\,\textminus 25.7\%)
\end{itemize}

\subsection{Performance Under Field Conditions}
\begin{itemize}
\item \textbf{Low Texture:} AUROC 0.867, FPR 0.112
\item \textbf{High Contrast:} AUROC 0.912, FPR 0.067
\item \textbf{Shadow-Dense:} AUROC 0.843, FPR 0.134
\item \textbf{Far Field:} AUROC 0.789, FPR 0.198
\end{itemize}

\subsection{Hardware Performance Profile}
Real-time constraints are met; fusion of advanced anomaly maps increases memory usage but is guarded by profile-aware fallbacks.

\subsection{Detailed Performance Summaries}
Depth performance details:
\begin{itemize}
\item \textbf{RAE:} 0.156 (ARTPS) vs 0.234 (baseline)
\item \textbf{RMSE:} 0.189 (ARTPS) vs 0.287 (baseline)
\item \textbf{Threshold Accuracy:} 89.4\% ($\delta<1.25$) vs 76.8\% (baseline)
\end{itemize}
Curiosity ranking details:
\begin{itemize}
\item \textbf{nDCG@10:} 0.912 (ARTPS) vs 0.734 (baseline)
\item \textbf{Spearman:} 0.847 vs 0.623
\item \textbf{Kendall's Tau:} 0.689 vs 0.456
\end{itemize}

\section{Limitations and Future Work}
Failure cases include extreme low-texture expanses and overexposed regions. Future efforts target tighter uncertainty integration into the curiosity score, broader multimodal fusion (e.g., spectral/radar), and active exploration policies on edge devices with strict power budgets.

\section{Safety, Reliability, and Operations}
\subsection{Security and Fault Tolerance}
We implement input validation, bounds checking, and watchdog timers for long-running kernels. Conservative thresholds and hysteresis are applied to reduce oscillations under noisy conditions. Fallback modes degrade to core components when advanced cues are unavailable.

\subsection{Reliability and Test Strategy}
Unit and integration tests cover preprocessing, depth estimation interfaces, fusion, localization, and scoring. Continuous integration enforces deterministic seeds and report generation. \textbf{Code coverage:} core modules (enhancement, fusion, scoring) exceed 85\% statement coverage; data-path glue exceeds 70\% with emphasis on edge cases.

\subsection{Operational Suitability and Field Readiness}
We validate performance across temperature and illumination variations and simulate telemetry latency. CCSDS-compliant telemetry packets and image compression profiles are considered. Power-aware scheduling and batch sizing ensure operation within energy budgets.

\section{Implementation Details}
\subsection{Software Architecture and Technical Details}
The system is modular with clear interfaces between preprocessing, depth estimation, fusion, localization, and scoring. PyTorch modules are wrapped with deterministic evaluation paths, and OpenCV routines are isolated behind utility functions.

\subsection{Performance Optimization}
We use operator fusion where possible, minimize host–device transfers, and apply mixed precision selectively. Memory usage is reduced via in-place operations and tiling for high-resolution frames.

\subsection{User Interface and Experience}
Streamlit provides a demonstration UI exposing parameters (weights, thresholds) and diagnostic panels. \textbf{Note:} Streamlit is a demo; the main application runs locally (headless) for performance and edge constraints.

\subsection{Deployment and Installation}
We provide environment files and scripts for CPU/GPU deployment. The pipeline can run on x86 workstations or embedded GPUs with adjusted batch sizes and precision.

\subsection{Reproducibility and Open Science}
The project is released under the MIT License with documentation, configuration files, and scripts for end-to-end reproduction. We include data acquisition instructions and checkpoints where licensing permits.

\section{Conclusions}
\subsection{Key Contributions and Achievements}
ARTPS integrates single-image depth with multi-component anomaly fusion and a learnable curiosity score, improving AUROC/AUPRC and ranking quality while maintaining efficiency.

\subsection{Scientific and Technological Impact}
The approach advances autonomous scientific exploration by balancing novelty and known value and providing explainable diagnostics suitable for operator-in-the-loop workflows.

\subsection{Industrial Application Potential}
Beyond space missions, the pipeline applies to road-surface anomaly detection, industrial inspection, environmental monitoring, and medical imaging scenarios with limited bandwidth and compute.

\subsection{Future Directions}
We aim to integrate uncertainty more tightly into scoring, extend to multimodal fusion, and develop active exploration policies for edge compute.

\subsection{Final Assessment}
ARTPS offers a practical path toward more autonomous, reliable, and explainable prioritization in planetary exploration under real operational constraints.

\end{document}